\documentclass[sigconf, preprint]{acmart}
\AtBeginDocument{%
  }

\setcopyright{acmlicensed}
\copyrightyear{2018}
\acmYear{2018}
\acmDOI{XXXXXXX.XXXXXXX}
\acmConference[Conference acronym 'XX]{Make sure to enter the correct
  conference title from your rights confirmation email}{June 03--05,
  2018}{Woodstock, NY}

\renewcommand\footnotetextcopyrightpermission[1]{}
\acmISBN{978-1-4503-XXXX-X/2018/06}

\usepackage{booktabs} 

\usepackage{graphicx}
\usepackage{subfiles}
\usepackage{array}
\usepackage{amsmath}
\usepackage{algpseudocode}
\usepackage{hyperref}
\usepackage{multirow}
\usepackage{subfigure}
\usepackage{caption}
\usepackage{flushend}
\usepackage{balance}
\usepackage{diagbox}
\usepackage{colortbl}
 \usepackage{enumitem}
 \usepackage{balance}
 \usepackage{makecell}
 \usepackage{float}
\usepackage{fancyhdr}
\usepackage{threeparttable}

\usepackage{url}
\usepackage{bm}
\usepackage{comment}
\usepackage{graphicx}
\usepackage{pifont}
\usepackage{tcolorbox}

\usepackage{subfigure}

\usepackage[ruled,linesnumbered]{algorithm2e}

\usepackage{listings}

\newcommand{\tool}{\textsc{PatMD}}

\begin{document}

\title{Fall into a Pit, Gain in a Wit‌: Cognitive-Guided Harmful Meme Detection via Misjudgment Risk Pattern Retrieval}

\author{Wenshuo Wang$^{1,2,3,\dagger}$, Ziyou Jiang$^{1,2,3,\dagger}$, Junjie Wang$^{1,2,3{*}}$, Mingyang Li$^{1,2,3}$, Jie Huang$^{1,2,3}$, Yuekai Huang$^{1,2,3}$, Zhiyuan Chang$^{1,2,3}$, Feiyan Duan$^{1,2,3}$, and Qing Wang$^{1,2,3}$}

\affiliation{$^{1}$State Key Laboratory of Complex System Modeling and Simulation Technology, Beijing, China;\\
$^{2}$Science and Technology on Integrated Information System Laboratory, \\
Institute of Software Chinese Academy of Sciences, Beijing, China;\\
$^{3}$University of Chinese Academy of Sciences, Beijing, China.\\
\{wangwenshuo2024, ziyou2019, junjie, mingyang2017, huangjie, \\yuekai2018, zhiyuan2020, duanfeiyan2025, wq\}@iscas.ac.cn\\
\country{}}

\authornote{Corresponding author.\\$\dagger$ These authors contributed equally to this work. }

\renewcommand{\shortauthors}{Wang et al.}

\begin{abstract}

Internet memes have emerged as a popular multimodal medium, yet they are increasingly weaponized to convey harmful opinions through subtle rhetorical devices like irony and metaphor. 
Existing detection approaches, including Multimodal Large Language Model (MLLM)-based techniques, struggle with these implicit expressions, leading to frequent misjudgments.
This paper introduces {\tool}, a novel approach that detects harmful memes by learning from and proactively mitigating these potential misjudgment risks. 
Our core idea is to move beyond superficial content-level matching and instead identify the underlying misjudgment risk patterns, proactively guiding the MLLMs to avoid known misjudgment pitfalls. 
We first construct a knowledge base where each meme is deconstructed into a misjudgment risk pattern explaining why it might be misjudged, either overlooking harmful undertones (false negative) or overinterpreting benign content (false positive). 
For a given target meme, {\tool} retrieves relevant patterns and utilizes them to dynamically guide the MLLM's reasoning.
Experiments on a benchmark of 6,626
memes across 5 harmful detection tasks show that {\tool} outperforms state-of-the-art baselines, achieving an average of 8.30\% improvement in F1-score and 7.71\% improvement in accuracy, while exhibiting consistent robustness on unseen and adversarial memes.

\noindent \textcolor{red}{Disclaimer: This paper contains content for demonstration purposes that may be disturbing to some readers.}
\end{abstract}



\maketitle

\section{Introduction}

{Nowadays, Internet memes have emerged as one of the most vibrant cultural phenomena of the digital age. 
They combine images and text, leveraging their humorous and satirical content to express viewpoints, fostering community cohesion, and even influencing public opinion~\cite{lin2024goat}.
However, this powerful influence is a double-edged sword, because the widespread nature of memes makes them a vehicle for amplifying hateful speech, reinforcing biases, and disseminating discrimination~\cite{sharma2022detecting}, posing a threat to online platforms.}

\begin{figure}[t]
\centering
\includegraphics[width=\columnwidth]{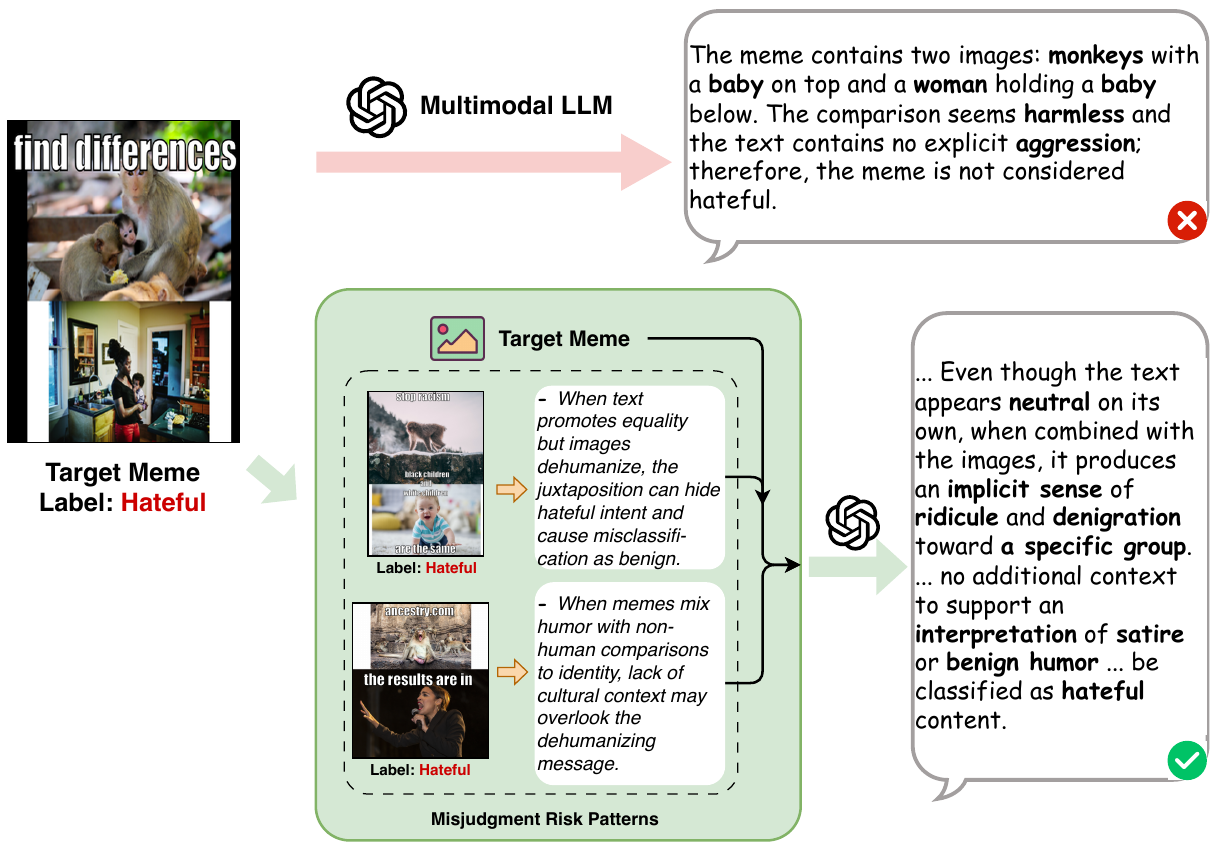}
\caption{The motivation example of harmful meme detection with misjudgment risk patterns.}
\label{fig:motivation_meme_detection}
\end{figure}

To detect harmful memes and safeguard online platforms, previous research has evolved from feature-fusion methods~\cite{suryawanshi2020offensive, pramanick-etal-2021-momenta-multimodal, lu2019vilbert, chen2020uniter} to MLLM-based approaches~\cite{hee2022explaining, lin-etal-2023-beneath, lin2025ask}, including recent Retrieval-Augmented Generation (RAG) paradigms~\cite{huang2024towards, liu2025mind} that reference historical memes to assist decision-making. 
However, these advanced methods remain fundamentally \textit{content-centric}. They primarily rely on superficial visual and textual similarity, which makes them struggle when novel memes employ entirely different templates to convey similar implicit rhetoric (e.g., sarcasm or metaphor)~\cite{kiela2020hateful, lin2024goat}. More importantly, these prior works fail to formalize the MLLM's inherent reasoning pitfalls into generalized and reusable misjudgment risk pattern that explicitly explain the underlying reasons for potential misjudgments.

{Figure \ref{fig:motivation_meme_detection} illustrates the contribution of cognitive blind spot. The upper part of MLLM's response shows that while standard MLLM can identify isolated elements (e.g., a human and a monkey), they fail to capture \textbf{why and how their juxtaposition constitutes a racially hateful analogy}. 
In contrast, our approach introduces a generalized \textit{misjudgment risk pattern} that incorporates the human cognitive and decision-making process in analyzing harmful memes, proactively warning the model of derogatory human-animal comparisons intended to deliberate dehumanization. Equipped with this structured thought, the MLLM transcends isolated visual features to comprehend the implicit rhetorical relations, successfully mitigating the misjudgment and identifying the harmful meme.}

Based on this thought of misjudgment risk pattern, our work presents a paradigm shift from purely \textit{\textbf{content-centric}} detection to a \textit{\textbf{cognitive decision-centric}} approach. Our core idea is to preemptively abstract systematic failure modes into \textit{misjudgment risk patterns}. By explicitly cautioning the MLLM against common cognitive pitfalls (e.g., over-interpreting irony or missing analogies) during its reasoning, this approach naturally fosters a more transparent decision logic.
However, transitioning to this paradigm entails two key challenges: (1) Historical misjudgments are scattered and unstructured, necessitating a systematic framework to categorize and archive these cognitive blind spots; (2) There lacks an effective mechanism to extract these implicit reasoning patterns from past errors and accurately retrieve them for novel memes.

In this paper, we propose \textbf{\tool}, an automated approach that leverages misjudgment risk \textbf{Pat}terns to improve harmful \textbf{M}eme \textbf{D}etection. 
Inspired by the concept ``\textbf{\textit{Fall into a Pit, Gain in a Wit}}'', 
where ``\textbf{\textit{Pit}}'' refers to the inherent misjudgment traps that models often encounter when processing memes, and ``\textbf{\textit{Wit}}'' represents the derived cognitive patterns used to steer the MLLM away from these pitfalls.
{\tool} first constructs a knowledge base by diagnostically decomposing memes to archive past MLLM misjudgments.  
For any target meme, it retrieves these misjudgment risk patterns and injects them into the MLLM's reasoning process, steering the unknown analysis away from known pitfalls: 

In \textit{misjudgment risk pattern elicitation stage}, historical memes are mapped into a hierarchical harm tree, capturing multi-granular harmful concepts, from which corresponding misjudgment risk patterns are derived. 
Then \textit{risk-aware pattern retrieval stage} identifies semantically similar memes based on structural similarity within the hierarchical harm tree, and fetches their associated risk patterns.
In \textit{pattern-augmented reasoning for the detection stage,} the retrieved patterns are integrated into a structured prompt, guiding the MLLM to avoid known misjudgment pitfalls. 

To evaluate the performance of {\tool}, we conduct extensive experiments on a benchmark with 6,626 memes spanning 5 harmful detection tasks. 
The results demonstrate that {\tool} has substantial and consistent improvements against four specialized baselines and six general-purpose MLLMs, with an average improvement of 8.30\% in F1-score and 7.71\% in accuracy. 
The ablation study reveals the necessity of each component in our approach design.
Moreover, {\tool} demonstrates exceptional generality and robustness, effectively adapting to unseen domains and resisting adversarial visual perturbations.
Further analysis reveals that the strength of {\tool} lies in its ability to steer MLLM toward proactive misjudgment prevention by focusing on the underlying logic of harm manifestation, rather than relying on superficial content analysis. 
This misjudgment-aware reasoning paradigm enables the model to perform more accurate and nuanced judgments, effectively reducing prevalent misclassification errors.

The main contributions of this paper are summarized as follows:
{
\begin{itemize}[leftmargin=*]
    \item {We propose {\tool}, a cognitive-guided approach that improves harmful meme detection by learning from misjudgment risk patterns of historical memes, which proactively steers the MLLM's reasoning away from misjudgment pitfalls.}

    \item We conduct experiments on a benchmark covering five harmful detection tasks. The results demonstrate that {\tool} outperforms baselines on F1-score with high consistent robustness against unseen and adversarial memes.

\end{itemize}
}
\section{Related Works}

Harmful meme detection has developed into an important direction in multimodal research and has continued to advance with the introduction of large-scale benchmark datasets~\cite{cai2019sarcasm, kiela2020hateful, fersini2022mami, lin2024goat}.
Early studies mainly focused on feature fusion, typically adopting a two-stream architecture that extracts visual and textual representations through independent encoders and then leverages attention mechanisms for cross-modal interaction and alignment~\cite{suryawanshi2020offensive, pramanick-etal-2021-momenta-multimodal, lee2021DisMultiHate}. Subsequently, research shifted towards shared Transformer-based intermediate fusion and fine-tuning of pretrained multimodal models, both of which jointly model image regions and text sequences to enhance cross-modal alignment and discrimination~\cite{kiela2019supervised, lu2019vilbert, li2019visualbert, chen2020uniter}. With the introduction of the Hateful Memes Challenge~\cite{kiela2020hateful}, a variety of ensemble-based and contrastive reweighting approaches emerged~\cite{muennighoff2020vilio, sandulescu2020detecting, lippe2020multimodal}, further improving their generalizability.
Other researchers explore prompt-based methods. Some work in this area~\cite{cao2022prompting, Creative-Harm-23-Ji, cao2023pro, ji2024capalign} focused on concatenating the meme's text with extracted image captions to fine-tune pre-trained language models\cite{liu2019roberta, chung2024scaling} for the detection task. 

\begin{figure*}[t]
\centering
\includegraphics[width=\linewidth]{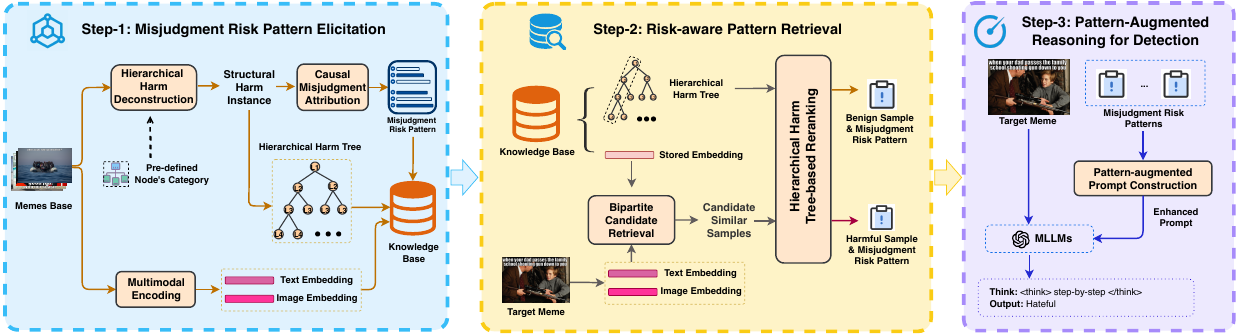}

\Description{A flow diagram showing the modules of our proposed method, 
including construct, retrieve, and inference.}
\caption{The overview of {\tool}.}
\label{fig:method}
\end{figure*}

Recently, MLLMs have demonstrated exceptional capabilities in complex reasoning~\cite{brown2020language, chowdhery2023palm}, such as generating intermediate reasoning steps via Chain-of-Thought prompting to enhance judgment accuracy~\cite{kojima2022large, wei2022chain, zhang2022automatic}. Inspired by this, research in interpretable harmful meme detection has begun to leverage LLMs to generate explicit chains of reasoning to explain their rationale. These methods, such as constructing LLM debate frameworks \cite{hee2022explaining} or distilling LLM's reasoning knowledge into smaller models \cite{lin-etal-2023-beneath}. Building upon this interactive paradigm, recent work~\cite{lin2025ask} has employed a multi-agent framework that simulates a discussion where agents query a vision expert for deeper visual understanding.
To address challenges in low-resource generalization, researchers have explored few-shot enhanced methods~\cite{cao2024modularized, huang2024towards}. Inspired by this direction, Liu et al. introduced MIND~\cite{liu2025mind}, a framework for zero-shot harmful meme detection that integrates similar example retrieval and multi-agent collaboration.
This aligns with the broader paradigm of RAG~\cite{lewis2020retrieval, guu2020retrieval}, which enriches an LLM's input with external knowledge and has proven highly effective in knowledge-intensive tasks~\cite{ram2023context, jiang2023active, asai2024self, wei2024instructrag}.

While the aforementioned approaches improve performance by enhancing content understanding or providing content examples, they remain fundamentally content-centric. When malicious users craft novel and implicit memes, these methods may struggle to find relevant content for reference. Inspired by these lines of research, we propose a paradigm shift. Instead of focusing on ``what the content is," our method focuses on ``why the model might misjudge." We introduce an approach based on misjudgment risk patterns and integrate it into the LLM's decision-making process. This cognitive decision-centric perspective allows it to identify and guard against specific reasoning paths that lead to errors.

\section{Overview of {\tool}}

Figure \ref{fig:method} illustrates the structure of {\tool}, which formalizes the flow ``Memes Base $\rightarrow$ Pattern Elicitation $\rightarrow$ Pattern Retrieval $\rightarrow$ Detection'', to detect harmful memes by integrating MLLM's capabilities within a RAG paradigm.
We shift the paradigm from reliance on the MLLM's implicit knowledge to proactively guide it with explicit misjudgment risk patterns derived from historical memes.
These patterns capture the specific manner in which the MLLM may misinterpret the multimodal content. During inference, relevant patterns are used to steer the MLLM's reasoning process, proactively alerting it to potential pitfalls and thereby enabling more robust judgments.
The implementation of {\tool} is structured into three main stages:

\textbf{Misjudgment Risk Pattern Elicitation}: 
We deconstruct historical memes into a hierarchical harm tree, which structurally represents harmful concepts through a macro-to-micro schema. This multi-level representation then serves as the foundation for generating the meme's corresponding misjudgment risk pattern, which captures how its semantic and contextual composition may trigger specific reasoning failures in MLLMs.

\textbf{Risk-aware Pattern Retrieval}: For a target meme, this stage performs a two-step, risk-aware retrieval process. After gathering candidate memes based on coarse content similarity, it then reranks them using a hierarchical harm tree-based similarity measure to identify those with the most analogous harmful composition. 

\textbf{Pattern-Augmented Reasoning for Detection}: Finally, the patterns from the retrieved memes are dynamically integrated into a structured prompt. This guides the MLLM to perform a well-reasoned analysis, safeguarding it against the misjudgments identified in the historical patterns.

\subsection{Misjudgment Risk Pattern Elicitation}
\label{sec:base_construction}

This Misjudgment Risk Pattern Elicitation stage forms the cornerstone of our approach, which provides the patterns that are used to steer the MLLM's reasoning process.
This stage first conducts the hierarchical harm deconstruction to systematically capture the diversified views of the harmful concepts in a meme, followed by the causal misjudgment attribution, aiming at deriving the misjudgment risk pattern.

\subsubsection{\textbf{Hierarchical Harm Deconstruction}}\label{sec:hier_harm_deconstruction}

{To systematically capture the diversified view of harmful composition, we first perform the hierarchical harm deconstruction. 
This process employs a ``Macro to Micro'' schema to structurally dissect a meme's harmful composition, delineating its underlying harmful logic layer by layer and moving beyond superficial features to reveal its core harmful structure. 
The results across all memes collectively form a Hierarchical Harm Tree $\mathcal{T}_\text{H}$, where each meme corresponds to a distinct structural harm instance within the tree.

\noindent\textbf{Hierarchical Harm Tree Structure Definition.}
The tree incorporates a root node $RNode$ and other tree nodes $TNode$. 
If the tree node is at the bottom level, it is considered a leaf node $LNodes$. The tree can be formalized as follows:
\begin{equation}
    \mathcal{T}_\text{H}=
    \left\{
    \begin{array}{lr}
       RNode\rightarrow [TNode^{(1)}_{L_1},...,TNode^{(N)}_{L_1}]  &  \\
       TNode_{L_i}^{(j)}\rightarrow [TNode^{(1)}_{L_{i+1}},...,TNode^{(N_j)}_{L_{i+1}}],\\ \quad \text{where} \; i\in\{1,...,4\}\;\text{and}\;TNode_{L_4}^{(i)}=LNode\\
       LNode_{L_4}^{(i)}\rightarrow \emptyset
    \end{array}
    \right.
\end{equation}
where the symbol $\rightarrow$ is the direction of the tree's layer. For each tree node $TNode^{(j)}_{L_i}$, the subscript $L_i$ means the level of this node, and the superscript $(j)$ is the $j_{th}$ node of the $L_{i}$ level's node, and $N_j$ is the number of its child nodes.
{Based on the harm expressions in memes, and inspired by the framework for harmful category classification} in~\cite{chen2025adammeme}, we define a four-level tree $[L_1,...,L_4]$. The content of each layer's node is defined as follows:
\begin{itemize}[leftmargin=*]
    \item \textbf{$L_1$ (Macro Category):} Identifies the category of the content (i.e., Nationality, Gender, Religion, Race, Animal, Disability, Child Exploitation, Political, None)
    \item \textbf{$L_2$ (Core Subjects):} Recognizes the specific groups, symbols, or figures explicitly targeted in the meme.
    \item \textbf{$L_3$ (Expressive Techniques):} Analyzes the key methods used by the creator to convey the message (e.g., use of stereotypes, irony, direct assertion).
    \item \textbf{$L_4$ (Instantiation of Techniques):} Describes the concrete manifestation of the aforementioned techniques in the meme.
\end{itemize}

\noindent\textbf{Automated Tree Construction.}
{Guided by the defined hierarchical structure, we employ an MLLM to automatically decompose every meme's content into 
conceptual nodes at each level of the hierarchical harm tree.}

Specifically, for a given meme $m_i$, the generation function $\text{MLLM}_{\text{gen}}$ fills the nodes of the tree template $\text{Template}(\mathcal{T}_\text{H})$ to produce the final instantiated tree, formulated as $\mathcal{T}_\text{H}^{(i)} = \text{MLLM}_{\text{gen}}(m_i, \text{Template}(\mathcal{T}_\text{H}))$ (See Appendix~\ref{sec:appendix_prompt} Figure~\ref{fig:prompt_tree} for the detailed prompt templates).
This process yields a structural harm instance in the hierarchical harm tree.
For instance, as shown in Figure \ref{fig:tree_construct}, the MLLM would identify the `Macro Category' ($L_1$) as ``Disability'', the `Core Subject' ($L_2$) as ``Child in a wheelchair'', and so on. 

\begin{figure}[h]
\centering
\includegraphics[width=\columnwidth]{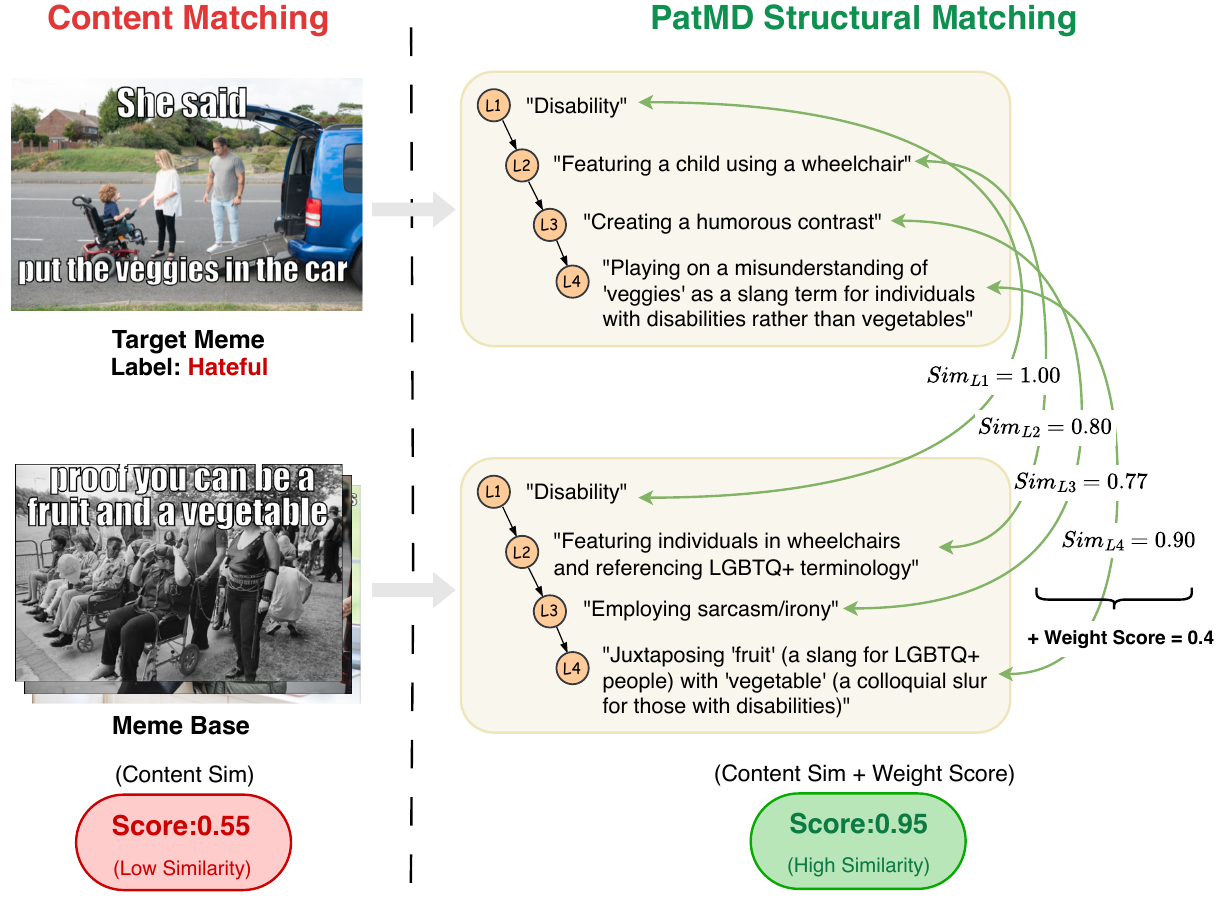}
\vspace{-0.7cm}
\caption{{The example of {\tool} hierarchical harm tree and structural matching versus content matching.}
}

\vspace{-0.7cm}
\label{fig:tree_construct}
\end{figure}

\subsubsection{\textbf{Causal Misjudgment Attribution}}\label{sec:causal_misjudg_att}

Building upon the deconstructed harm profile, we then derive the misjudgment risk pattern. 
We analyze how the identified harmful concepts could lead an MLLM to specific misjudgments (e.g., overlooking ironic cues or misinterpreting contextual references), and produce the misjudgment risk pattern for each meme.
We posit that the processes for generating misjudgment risk patterns differ fundamentally between benign and hateful samples. Therefore, {\tool} employs distinct analytical strategies for each:


\begin{itemize}[leftmargin=*]
    \item \textbf{For benign samples}, {\tool} infers the root cause of its \textit{false positive} risk based on the structural harm information. It analyzes why the combination of its theme, entities, and expressive techniques is misinterpreted or over-interpreted as harmful.
    \item \textbf{For harmful samples}, {\tool} reasons in reverse, determining how the meme's compositional elements are subtly employed to encode or conceal its \textit{false negative} risk, which is its underlying harmful intent.
\end{itemize}

{This dual-pathway analysis helps to address the possibility of both false positives and false negatives. 
The MLLM abstracts the harmful concepts into a concise, generalizable misjudgment risk pattern (See Appendix~\ref{sec:appendix_prompt} Figure~\ref{fig:prompt_pattern} for the detailed prompt templates).} 

{After the above process, each sample $m_i = (\text{img}_i, \text{txt}_i)$ is endowed with a structural harm instance $\mathcal{T}_i$ within the hierarchical harm tree and a misjudgment risk pattern $\mathcal{P}_i$. }
To make these information retrievable, we also employ a pre-trained multimodal encoder $E$~\cite{gunther2025jinaEmbeddings} to generate embedding vectors for its image and text, $v^{\text{img}}_i = E(\text{img}_i)$ and $v^{\text{txt}}_i = E(\text{txt}_i)$, respectively. 

{This stage ultimately produces a knowledge base where each meme is encoded as a tuple $(v^{\text{img}}_i, v^{\text{txt}}_i, \mathcal{T}_i, \mathcal{P}_i, \mathcal{G}_i)$, recording its multimodal embeddings, structural harm instance, misjudgment risk pattern, and ground truth label (benign/harmful) $\mathcal{G}_i$.}

\subsection{Risk-aware Pattern Retrieval}
\label{sec:retrieval_and_selection}

{To guide the follow-up detection of a target meme, {\tool} performs a risk-aware pattern retrieval. We go beyond content similarity by seeking memes that potentially share a comparable risk profile, as reflected by the hierarchical harm tree. {\tool} implements this by first gathering content-based candidates, then applying a hierarchical harm tree-based reranking to pinpoint the most structurally similar memes. }

\subsubsection{\textbf{Bipartite Candidate Retrieval}}
\label{sec:bipartite_retrieval}

The goal of the initial retrieval stage is to construct a diverse set of candidate samples.
We employ a label-based bipartite retrieval strategy to include both benign and harmful samples.

Specifically, we use the target sample's image vector $v_{\text{test}}^{\text{img}}$ and text vector $v_{\text{test}}^{\text{txt}}$ to separately query the subsets of samples corresponding to benign and harmful ground-truth labels. 
Within each labeled subset, we retrieve the top-$N$ most similar candidates from the image modality and the top-$N$ from the text modality. Subsequently, we take the union of all these retrieved candidates to form an initial candidate pool. This pool, containing samples that offer a balanced perspective on potential risks, is then passed to the reranking part for a more refined similarity assessment and reranking.

\subsubsection{\textbf{Hierarchical Harm Tree-based Reranking}}
\label{sec:reranking}

{The initial retrieval process, which relies on coarse vector similarity, may fail to adequately capture the nuanced manifestation of harmful concepts between the target and candidate memes.
We utilize the structural matching of the hierarchical harm tree to choose more  suitable memes.  }

We employ a top-down sequential matching mechanism that terminates at the first mismatched level. 
The matching sequence can start with a successful match at either the $L_1$ or $L_2$ level and proceed downwards. 
A match at the $L_1$ level requires an exact content match. 
For levels $L_2$ and below, a match is successful if the semantic similarity of their node descriptions exceeds a predefined threshold $\beta$.

For each successfully matched level ($L_1$, $L_2$, $L_3$, or $L_4$), a fixed bonus score $\gamma$ is added to the candidate's similarity score, formulated as follows:
\begin{equation}
\label{eq:tree_similarity_concise}
S_{\mathcal{T}} = \sum_{k=1}^{4} \gamma \prod_{i=1}^{k} M_i
\end{equation}
where $\gamma$ is the bonus for a match at level $L_k$. The match status $M_i \in \{0, 1\}$ is determined by:
\begin{itemize}[leftmargin=*]
    \item For $L_1$: $M_1=1$ if an exact content match occurs.
    \item For $L_{i>1}$: $M_i=1$ if the semantic similarity of node content exceeds a threshold $\beta$.
\end{itemize}
The product term $\prod_{i=1}^{k} M_i$ ensures that the bonus $\gamma$ is awarded only if all levels from $L_1$ to $L_k$ have successfully matched, thus enforcing the top-down sequential mechanism.
Consequently, the tree similarity score for a candidate, denoted as $S_{\mathcal{T}}$, is the hierarchical bonuses accumulated from all continuously matched levels (top-to-down).

Figure~\ref{fig:tree_construct} illustrates this advantage. Consider two hateful memes sharing the same derogatory logic (i.e., exploiting the slang ``vegetable’’). Due to distinct surface features, conventional content matching fails to link them, yielding a low similarity score of 0.55. Conversely, {\tool} hierarchically deconstructs them to capture implicit rhetorical analogies. By accumulating structural similarities across $L_1$--$L_4$ alongside hierarchical bonuses (+0.4 weight score), {\tool} bridges the semantic gap, boosting the final similarity score to 0.95 for accurate retrieval.

In addition, we also utilize a unified Multimodal Reranker\footnote{https://huggingface.co/jinaai/jina-reranker-m0} to refine the initial similarity scores. This component re-evaluates the visual relevance between the target and candidate images via attention-based interactions to generate a more reliable image similarity score, $S_{\text{img}}$, and performs semantic interaction on the target and candidate texts to output a more precise text similarity score, $S_{\text{txt}}$. 
Finally, we fuse the scores to compute a final similarity score for each candidate sample:
\begin{equation}
    S_{\text{final}} = \alpha \cdot S_{\text{img}} + (1-\alpha) \cdot S_{\text{txt}} + S_{\mathcal{T}}
    \label{eq:final_score}
\end{equation}
where $\alpha$ is a hyperparameter that balances the importance of the image and text modalities, and $S_{\mathcal{T}}$ represents the score derived from the structured harm matching. 

All candidates are sorted in descending order based on their $S_{\text{final}}$ score, and the top-$K$ samples are selected.
The corresponding misjudgment risk patterns of these selected samples are subsequently used to guide the follow-up detection.

\subsection{Pattern-Augmented Reasoning for Detection}
\label{sec:risk_detection}\label{sec:prompt_construction}

With the retrieved memes and their associated misjudgment risk patterns, we orchestrate them into a structured prompt and guide the MLLM through a systematic reasoning chain, steering it clear of historical pitfalls to arrive at the well-justified decision. 

We present pattern-augmented prompt construction, where the prompt is built upon a structured template designed to orchestrate the MLLM's analytical workflow. 
The template first grounds the MLLM by clearly stating the classification task and providing an objective definition of the target harm category adopted from~\cite{lin2025ask}. 
{Its core component is the \textit{Critical Reflection References} section, which contains the retrieved misjudgment patterns categorized based on their ground-truth labels.}

Patterns associated with harmful memes are aggregated and framed as \textit{Miss Reasons} to alert the MLLM to subtle harmful patterns it might otherwise overlook. 
Conversely, patterns from benign memes are presented as \textit{False Alarm Reasons} to prevent the MLLM from misclassifying non-hateful content. 
To ensure this guidance is used for critical reflection rather than simple imitation, the template also enforces a structured reasoning procedure that compels the model to form an independent analysis before consulting the provided support.
Through this strategy, a unique and highly relevant prompt is dynamically generated for each target sample (See Appendix~\ref{sec:appendix_prompt} Figure~\ref{fig:prompt_detection} for detailed prompts).

Finally, the complete prompt, populated with the target sample's content and the dynamically constructed \textit{Critical Reflection References}, is submitted to an MLLM.
Instructed to follow the explicit reasoning procedure, the MLLM performs a context-aware analysis. 
This guided process ensures that the final output is not merely a binary classification but a well-reasoned judgment, critically informed by relevant historical misjudgment patterns.

\section{Experiment Design}

{To evaluate the performance of {\tool}, we introduce the following two Research Questions (RQs).
\begin{itemize}[leftmargin=*]
    \item \textbf{RQ1: What are the performances of {\tool} on the harmful meme detection?} We aim to evaluate the performance of {\tool} against different baselines.
    \item \textbf{RQ2: What is the contribution of components to the harmful meme detection?} 
    {Since {\tool}'s framework follows the stages of the RAG paradigm, we aim to evaluate how {\tool} outperforms the original RAG's strategy, by comparing the performance of {\tool} and variants of removing components.} 
    {\item \textbf{RQ3: What is the performance of PatMD on unseen and adversarial memes? } 
    We aim to evaluate the generality and robustness of our approach {\tool} under out-of-distribution and perturbed scenarios.}

\end{itemize}}

\subsection{Datasets}

We use the GOAT-bench~\cite{lin2024goat}, a popular comprehensive benchmark of 6,626 memes for evaluating {harmful meme detection} across five tasks: hateful, misogynistic, offensive, sarcastic, and harmful content. This benchmark aggregates five specialized datasets (FHM~\cite{kiela2020hateful}, MAMI~\cite{fersini2022mami}, MultiOFF~\cite{suryawanshi2020offensive}, MSD~\cite{cai2019sarcasm}, Harm-C/P~\cite{pramanick-etal-2021-momenta-multimodal}). We utilize the train and test splits provided by GOAT-bench. 
Detailed dataset statistics are available in Table~\ref{tab:dataset}.

\begin{table}[t]
\caption{Statistics of memes that are used in {\tool}.}
\vspace{-0.2cm}
\centering
\resizebox{\linewidth}{!}{%
\begin{tabular}{c|ccccc}
\toprule
\textbf{Type} & \textbf{Hatefulness} & \textbf{Misogyny} & \textbf{Offensiveness} & \textbf{Sarcasm} & \textbf{Harmfulness} \\ \midrule
Train & 8,500 & 10,000 & 7,000 & 19,816 & 4,250 \\
Test & 2,000 & 1,000 & 743 & 1,820 & 1,063 \\ \bottomrule
\end{tabular}%
}
\vspace{-0.5cm}
\label{tab:dataset}
\end{table}

\begin{table*}[t]
\centering
\caption{Performance of our proposed {\tool} and the baselines across different tasks (\%).}

\resizebox{\textwidth}{!}{%
\begin{tabular}{l|cccccccccc|cc}
\toprule
\multicolumn{1}{c|}{}                                 & \multicolumn{2}{c}{\textbf{Hatefulness}} & \multicolumn{2}{c}{\textbf{Misogyny}} & \multicolumn{2}{c}{\textbf{Offensiveness}} & \multicolumn{2}{c}{\textbf{Sarcasm}} & \multicolumn{2}{c|}{\textbf{Harmfulness}} & \multicolumn{2}{c}{\textbf{\textit{Overall}}} \\ 
\multicolumn{1}{c|}{\multirow{-2}{*}{\textbf{Approach}}} & Acc                 & F1                 & Acc               & F1                & Acc                      & F1              & Acc               & F1               & Acc                 & F1                 & Acc               & F1               \\
\midrule
\multicolumn{13}{c}{\textbf{\textit{Specialized Harmful Meme Detection Baselines}}}
\\
\midrule
PromptHarm~\cite{cao2022prompting}                                           & 64.05               & 51.57              & 69.30             & 67.81             & 61.88                    & 57.89           & 58.51             & 55.08            & 61.71               & 53.44              & 63.28             & 57.16            \\
Pro-Cap~\cite{cao2023pro}                                              & 52.40               & 51.80              & 61.56             & 58.57             & 48.56                    & 47.15           & 54.90             & 54.62            & 57.06               & 56.83              & 54.90             & 53.79            \\
CapAlign~\cite{ji2024capalign}                                             & 63.70               & 51.60              & 68.60             & 66.81             & 61.51                    & 54.76           & 58.08             & 52.67            & 60.77               & 51.04              & 62.53             & 55.38            \\
MemeAgent~\cite{lin2025ask}                             & 68.40               & 62.73              & 73.10             & 73.04             & 62.18                    & 58.96           & 70.11             & 69.97            & 63.23               & 61.25              & 67.40             & 65.19            \\
\rowcolor[HTML]{ECF4FF} 
$\Delta$(GPT-4o+{\tool}, SOTA)                              & $\textcolor[HTML]{990066}{+12.55}$                & $\textcolor[HTML]{990066}{+16.41}$                & $\textcolor[HTML]{990066}{+13.10}$                & $\textcolor[HTML]{990066}{+13.14}$                & $\textcolor[HTML]{990066}{+2.64}$                    & $\textcolor[HTML]{003300}{-0.02}$                & $\textcolor[HTML]{990066}{+12.42}$                & $\textcolor[HTML]{990066}{+12.44}$                & $\textcolor[HTML]{990066}{+5.35}$                 & $\textcolor[HTML]{990066}{+5.44}$                 & $\textcolor[HTML]{990066}{+9.22}$                 & $\textcolor[HTML]{990066}{+9.48}$                 \\
\midrule
\multicolumn{13}{c}{\textbf{\textit{MLLM-based Baselines and Our Approach}}}
\\
\midrule
Qwen2.5VL-32B                                        & 72.85               & 68.25              & 65.20             & 60.89             & 63.17                    & 55.33           & 78.24             & 77.95            & 62.26               & 53.88              & 68.34             & 63.26            \\
\rowcolor[HTML]{ECF4FF} 
Qwen2.5VL-32B + \tool                 & \textbf{74.55}      & \textbf{72.19}     & \textbf{78.9}     & \textbf{78.84}    & 63.06                    & \textbf{60.86}  & \textbf{81.15}    & \textbf{81.15}   & \textbf{65.25}      & \textbf{63.77}     & \textbf{72.58}    & \textbf{71.36}   \\
\rowcolor[HTML]{ECF4FF} 
$\Delta$({\tool}, MLLM)                       & $\textcolor[HTML]{990066}{+1.70}$                & $\textcolor[HTML]{990066}{+3.94}$               & $\textcolor[HTML]{990066}{+13.70}$             & $\textcolor[HTML]{990066}{+17.95}$             & $\textcolor[HTML]{003300}{-0.11}$                    & $\textcolor[HTML]{990066}{+5.53}$            & $\textcolor[HTML]{990066}{+2.91}$              & $\textcolor[HTML]{990066}{+3.20}$             & $\textcolor[HTML]{990066}{+2.99}$                & $\textcolor[HTML]{990066}{+9.89}$               & $\textcolor[HTML]{990066}{+4.24}$              & $\textcolor[HTML]{990066}{+8.10}$             \\ \hline
InternVL3.5-20B-A4B                                 & 74.20               & 71.76              & 76.10             & 75.86             & 61.37                    & 54.09           & 66.15             & 62.59            & 59.17               & 52.78              & 67.40             & 63.42            \\
\rowcolor[HTML]{ECF4FF} 
InternVL3.5-20B-A4B + \tool          & \textit{72.8}       & \textit{69.15}     & \textbf{79.1}     & \textbf{79.03}    & \textit{\textbf{61.86}}  & \textit{52.38}  & \textbf{75.55}    & \textbf{75.47}   & \textbf{63.65}      & \textbf{58.5}      & \textbf{70.59}    & \textbf{66.91}   \\
\rowcolor[HTML]{ECF4FF} 
$\Delta$({\tool}, MLLM) 
& $\textcolor[HTML]{003300}{-1.40}$ & $\textcolor[HTML]{003300}{-2.61}$ & $\textcolor[HTML]{990066}{+3.00}$ & $\textcolor[HTML]{990066}{+3.17}$ & $\textcolor[HTML]{990066}{+0.49}$ & $\textcolor[HTML]{003300}{-1.71}$ & $\textcolor[HTML]{990066}{+9.40}$ & $\textcolor[HTML]{990066}{+12.88}$ & $\textcolor[HTML]{990066}{+4.48}$ & $\textcolor[HTML]{990066}{+5.72}$ & $\textcolor[HTML]{990066}{+3.19}$ & $\textcolor[HTML]{990066}{+3.49}$
             \\ \hline
InternVL3-14B                                        & 68.25               & 68.24              & 65.90             & 61.82             & 59.08                    & 59.06           & 58.68             & 50.34            & 57.86               & 57.75              & 61.95             & 59.44            \\
\rowcolor[HTML]{ECF4FF} 
InternVL3-14B + \tool                 & \textbf{76.95}      & \textbf{75.06}     & \textbf{77.80}    & \textbf{77.60}    & \textbf{63.26}           & 57.18           & \textbf{80.03}    & \textbf{79.88}   & \textbf{66.70}      & \textbf{64.99}     & \textbf{72.95}    & \textbf{70.94}   \\
\rowcolor[HTML]{ECF4FF} 
$\Delta$({\tool}, MLLM)                       & $\textcolor[HTML]{990066}{+8.70}$   & $\textcolor[HTML]{990066}{+6.82}$  & $\textcolor[HTML]{990066}{+11.90}$ & $\textcolor[HTML]{990066}{+15.78}$ & $\textcolor[HTML]{990066}{+4.18}$ & $\textcolor[HTML]{003300}{-1.88}$ & $\textcolor[HTML]{990066}{+21.35}$ & $\textcolor[HTML]{990066}{+29.54}$ & $\textcolor[HTML]{990066}{+8.84}$ & $\textcolor[HTML]{990066}{+7.24}$ & $\textcolor[HTML]{990066}{+10.99}$ & $\textcolor[HTML]{990066}{+11.50}$
            \\ \hline
LLaVA-1.6-13B                                        & 57.10               & 56.98              & 51.80             & 37.68             & 46.03                    & 40.09           & 50.00             & 33.33            & 52.30               & 49.65              & 51.45             & 43.54            \\
\rowcolor[HTML]{ECF4FF} 
LLaVA-1.6-13B + \tool                 & \textbf{64.47}      & 52.58              & \textbf{63.32}    & \textbf{58.27}    & \textbf{60.62}           & \textbf{46.20}  & \textbf{59.92}    & \textbf{54.03}   & \textbf{58.44}      & 45.88              & \textbf{61.35}    & \textbf{51.39}   \\
\rowcolor[HTML]{ECF4FF} 
$\Delta$({\tool}, MLLM)                       & $\textcolor[HTML]{990066}{+7.37}$                & $\textcolor[HTML]{003300}{-4.40}$              & $\textcolor[HTML]{990066}{+11.52}$             & $\textcolor[HTML]{990066}{+20.59}$             & $\textcolor[HTML]{990066}{+14.59}$                    & $\textcolor[HTML]{990066}{+6.11}$            & $\textcolor[HTML]{990066}{+9.92}$              & $\textcolor[HTML]{990066}{+20.70}$            & $\textcolor[HTML]{990066}{+6.14}$                & $\textcolor[HTML]{003300}{-3.77}$              & $\textcolor[HTML]{990066}{+9.91}$              & $\textcolor[HTML]{990066}{+7.85}$             \\ \hline
GPT-4o                                               & 76.60               & 75.24              & 82.71             & 82.43             & 59.16                    & 59.13           & 68.85             & 66.27            & 66.32               & 64.09              & 70.73             & 69.43            \\
\rowcolor[HTML]{ECF4FF} 
GPT-4o + \tool                        & \textbf{80.95}      & \textbf{79.14}     & \textbf{86.2}     & \textbf{86.18}    & \textbf{64.82}           & 58.94  & \textbf{82.53}    & \textbf{82.41}   & \textbf{68.58}      & \textbf{66.69}     & \textbf{76.62}    & \textbf{74.67}   \\
\rowcolor[HTML]{ECF4FF} 
$\Delta$({\tool}, MLLM) &                      $\textcolor[HTML]{990066}{+4.35}$   & $\textcolor[HTML]{990066}{+3.90}$  & $\textcolor[HTML]{990066}{+3.49}$  & $\textcolor[HTML]{990066}{+3.75}$  & $\textcolor[HTML]{990066}{+5.66}$ & $\textcolor[HTML]{003300}{-0.19}$ & $\textcolor[HTML]{990066}{+13.68}$ & $\textcolor[HTML]{990066}{+16.14}$ & $\textcolor[HTML]{990066}{+2.26}$ & $\textcolor[HTML]{990066}{+2.60}$ & $\textcolor[HTML]{990066}{+5.89}$  & $\textcolor[HTML]{990066}{+5.24}$ 
             \\ \hline
GPT-4o-mini                                          & 68.85               & 66.41              & 79.30             & 78.92             & 57.95                    & 57.82           & 57.97             & 49.29            & 66.50               & 66.10              & 66.11             & 63.71            \\
\rowcolor[HTML]{ECF4FF} 
GPT-4o-mini + \tool                   & \textbf{71.41}      & \textbf{66.97}     & 78.70              & 78.30              & \textbf{64.62}           & \textbf{62.47}  & \textbf{80.27}    & \textbf{80.10}    & \textbf{68.02}      & \textbf{67.64}     & \textbf{72.60}    & \textbf{71.10}   \\
\rowcolor[HTML]{ECF4FF} 
$\Delta$({\tool}, MLLM)                       & $\textcolor[HTML]{990066}{+2.56}$   & $\textcolor[HTML]{990066}{+0.56}$  & $\textcolor[HTML]{003300}{-0.60}$ & $\textcolor[HTML]{003300}{-0.62}$ & $\textcolor[HTML]{990066}{+6.67}$ & $\textcolor[HTML]{990066}{+4.65}$   & $\textcolor[HTML]{990066}{+22.30}$ & $\textcolor[HTML]{990066}{+30.81}$ & $\textcolor[HTML]{990066}{+1.52}$ & $\textcolor[HTML]{990066}{+1.54}$ & $\textcolor[HTML]{990066}{+6.49}$  & $\textcolor[HTML]{990066}{+7.39}$ 
             \\ 
\bottomrule

\end{tabular}
}
\label{tab:results}
\end{table*}



\subsection{Baselines}
We compare our approach against two categories of baselines: specialized approaches for harmful meme detection and general-purpose MLLMs.

\noindent\textbf{Specialized Approaches for Harmful Meme Detection.}

This category includes methods specifically designed or fine-tuned for this task. We consider two sub-types. The first are supervised methods that require task-specific training, including \textbf{PromptHarm}~\cite{cao2022prompting}, \textbf{Pro-Cap}~\cite{cao2023pro}, and \textbf{CapAlign}~\cite{ji2024capalign}. 
The second subtype comprises training-free frameworks, represented by \textbf{MemeAgent}~\cite{lin2025ask}, a SOTA multi-agent framework for meme analysis. 
We strictly adhere to the rigorous evaluation protocol and implementations established by the MemeAgent study to ensure a fair comparison.

\noindent\textbf{General-Purpose MLLMs.}
To demonstrate the effectiveness and generalizability of our approach, we establish baselines using a range of SOTA, general-purpose MLLMs. This includes \textbf{Qwen2.5-VL-32B-Instruct}~\cite{bai2025qwen2}, \textbf{InternVL3.5-20B-A4B}~\cite{wang2025internvl3.5}, \textbf{InternVL3-14B}~\cite{zhu2025internvl3}, \textbf{LLaVA-1.6-13B}\footnote{The baseline performance for LLaVA-1.6-13B is sourced from the MemeAgent~\cite{lin2025ask} study for consistency.} ~\cite{liu2024improved}, \textbf{GPT-4o}~\cite{hurst2024gpt}, and \textbf{GPT-4o-mini}~\cite{hurst2024gpt}.

\subsection{Metrics}
To ensure a comprehensive and impartial evaluation, we measure model performance using two core metrics: Accuracy and Macro-F1 Score. This choice aligns with the established evaluation protocol of the \textit{GOAT-Bench}, thereby enabling a direct and consistent comparison with prior work.

\subsection{Experimental Setup}

All experiments were conducted on a server equipped with a single NVIDIA H100 GPU with 80GB of VRAM. For the open-source models, we deployed local inference endpoints using the vLLM framework~\cite{kwon2023efficient}, loading pre-trained weights sourced from the HuggingFace Hub. For proprietary models, we accessed them via their official APIs, specifically using GPT-4o (\texttt{gpt-4o-2024-11-20} version) and GPT-4o-mini (\texttt{gpt-4o-mini-2024-07-18} version). Specifically, for the Misjudgment Risk Pattern Elicitation stage, we utilized GPT-4o as the MLLM.

To improve reproducibility, the temperature parameter was set to 0 during inference, which corresponds to deterministic decoding without sampling. In addition, the prompting strategy for the baseline strictly followed the configuration described in the previous work~\cite{lin2025ask} to guarantee fairness and consistency in comparison.

The hyperparameters for our method are set as follows: top-$N$=10, top-$K$=2, $\alpha=0.7$, $\beta=0.7$, and $\gamma=0.1$.

\section{Results and Analysis}

\begin{table*}[t]
\caption{The results of ablation studies between the {\tool} and RAG-oriented variants (\%).}

\vspace{-0.2cm}
\centering
\resizebox{\textwidth}{!}{
\begin{tabular}{l|cccccccccc|ll}
\toprule
\multicolumn{1}{c|}{\multirow{2}{*}{\textbf{Approach}}} & \multicolumn{2}{c}{\textbf{Hatefulness}} & \multicolumn{2}{c}{\textbf{Misogyny}} & \multicolumn{2}{c}{\textbf{Offensiveness}} & \multicolumn{2}{c}{\textbf{Sarcasm}} & \multicolumn{2}{c|}{\textbf{Harmfulness}} & \multicolumn{2}{c}{\textbf{\textit{Overall}}} \\ 
                                 & Acc                 & F1                 & Acc               & F1                & Acc                  & F1                  & Acc               & F1               & Acc                 & F1                 & \multicolumn{1}{c}{Acc}               & \multicolumn{1}{c}{F1}               \\ \midrule
InternVL3-14B + {\tool}                          & 76.95               & 75.06              & 77.80             & 77.60             & 63.26                & 57.18               & 80.03             & 79.88            & 66.70               & 64.99              & 72.95             & 70.94            \\
\hline
w/o Structural Reranking                         & 74.26               & 71.36              & 77.30             & 77.19             & 62.25                & 55.37               & 78.26             & 78.18            & 64.11               & 62.22              & 71.24\textsubscript{\textcolor[HTML]{003300}{(-1.71)}}             & 68.86\textsubscript{\textcolor[HTML]{003300}{(-2.08)}}           \\
w/o Semantic Reranking                       & 76.19               & 74.29              & 76.98             & 76.73             & 62.05                & 56.32               & 78.95             & 78.78            & 66.13               & 64.49              & 72.06\textsubscript{\textcolor[HTML]{003300}{(-0.89)}}            & 70.12\textsubscript{\textcolor[HTML]{003300}{(-0.82)}}            \\
Image-only RAG               & 75.70               & 73.74              & 77.34             & 77.29             & 62.48                & 56.39               & 78.04             & 77.89            & 64.35               & 62.26              & 71.58\textsubscript{\textcolor[HTML]{003300}{(-1.37)}}             & 69.51\textsubscript{\textcolor[HTML]{003300}{(-1.43)}}            \\
Text-only RAG               & 75.64               & 73.67              & 77.10             & 76.83             & 62.99                & 55.77               & 79.35             & 79.20            & 65.10               & 62.99              & 72.04\textsubscript{\textcolor[HTML]{003300}{(-0.91)}}             & 69.69\textsubscript{\textcolor[HTML]{003300}{(-1.25)}}            \\
Random RAG                 & 73.07               & 69.83              & 74.54             & 74.54             & 61.89                & 54.15               & 75.06             & 74.80            & 63.55               & 61.00              & 69.62\textsubscript{\textcolor[HTML]{003300}{(-3.33)}}            & 66.86\textsubscript{\textcolor[HTML]{003300}{(-4.08)}}            \\ \bottomrule
\end{tabular}
}
\vspace{-0.3cm}
\label{tab:ablation}
\end{table*}

\subsection{Performance of {\tool} (RQ1)}

To evaluate the effectiveness of our proposed approach, we compare the performance of {\tool} with the baselines and general-purpose MLLMs (i.e., MLLM in the table), as shown in Table~\ref{tab:results}.

\noindent\textbf{Overall Performance.}
The results demonstrate that {\tool} consistently improves the performance of specialized baselines and general-purpose MLLMs, with an average of 8.30\% improvement in F1-scores and an average of 7.71\% improvement in accuracy.
First, when compared against specialized harmful meme detection baselines, our method shows strong performance. For instance, when applied to GPT-4o and the open-source InternVL3-14B, our framework achieves overall F1-scores of 74.67\% and 70.94\%, respectively, both outperforming the strongest baseline, MemeAgent (65.19\%).
Second, the framework exhibits strong generalization across different MLLMs. It improves the overall F1-score of GPT-4o by 5.24\% and that of GPT-4o-mini by 7.39\%. The impact on open-source models is also notable. InternVL3-14B experiences an 11.50\% F1-score increase, surpassing the GPT-4o (70.94\% vs. 69.43\%). Similarly, LLaVA-1.6-13B, a model of comparable scale, receives an F1-score boost of 7.85\%. 
Furthermore, the framework proves effective even for specialized architectures such as InternVL3.5-20B-A4B, a Mixture-of-Experts (MoE) model activating only 4B parameters during inference, achieving an overall F1 improvement of 3.49\%.
These results validate that {\tool} is an effective and model-agnostic approach capable of enhancing the detection performance of MLLMs across different scales and architectures.

\noindent\textbf{Task-Specific Analysis.}
A detailed analysis of task-specific results further reveals the breadth and depth of the improvements. The framework's impact is particularly evident in Sarcasm and Misogyny detection. 
Notably, for Sarcasm, {\tool} yields substantial improvements across models. For instance, InternVL3-14B's F1-score improves by 29.54\%, and GPT-4o-mini's F1-score increases by 30.81\%. This highlights {\tool}'s ability to help models interpret the incongruity between text and image—a common characteristic of sarcastic memes.
In Misogyny detection, similar improvements are observed. Qwen2.5VL-32B's F1-score, for example, increases by 17.95\%. This indicates that by retrieving historical samples with similar stereotypical patterns, our framework effectively equips models to identify targeted, gender-based harm.

Beyond these tasks, {\tool} also demonstrates effectiveness across other domains. In Offensiveness detection, most models show positive gains, such as the 6.67\% F1-score increase for GPT-4o-mini. For Harmfulness, a general positive trend is also observed, with Qwen2.5VL-32B gaining 9.89\% in F1-score. These widespread gains underscore the framework's ability to provide task-relevant reasoning support.

\subsection{Ablation Study (RQ2)}\label{sec:ablation_study}

{To evaluate the effectiveness of the key components in our {\tool}, which covers all the RAG's framework in Figure \ref{fig:method}. We conduct a series of ablation studies, which evaluate the impact on {\tool}'s performance by removing or replacing key components (related to RAG paradigm) of our approach, including: the structural matching of hierarchical harm tree in Section \ref{sec:reranking} (\texttt{w/o Structural Reranking}), the image and text reranking in Section \ref{sec:reranking} (\texttt{w/o Semantic Reranking}), the text/image retrieval in Section \ref{sec:bipartite_retrieval} (\texttt{w/o Image/Text-only RAG}), and replacing the bipartite candidate retrieval of Section \ref{sec:bipartite_retrieval} with random retrieval (\texttt{Random RAG}).
}

As shown in Table~\ref{tab:ablation}, the complete {\tool} approach achieves the best performance in both accuracy and F1 score, demonstrating the integrity and synergistic effect of our design.
Among all variants, removing the structural matching of the hierarchical harm tree (\texttt{w/o Structural Reranking}) leads to one of the most significant performance drops, with the F1 score decreasing by 2.08\% (from 70.94\% to 68.86\%). This strongly demonstrates that our proposed hierarchical harm tree is crucial for the model to understand and judge the deep-seated risks in complex memes. 
Removing the image and text reranker (\texttt{w/o Semantic Reranking}), which is used to refine content similarity, also results in a performance decline. This indicates that a more precise assessment of content relevance is essential, besides performing the structure matching. 
Furthermore, performance drops when the text retrieval (\texttt{w/o Image-only RAG}) and image retrieval (\texttt{w/o Text-only RAG}) components are removed individually, validating the necessity of leveraging both modalities for retrieval, as relying on a single modality can miss key similar samples.
Finally, compared to random retrieval (\texttt{Random RAG}), our method shows a substantial performance advantage, with a F1 score more than 4\% higher. This highlights that accurately retrieving highly relevant historical memes from the knowledge base is vital to ensuring the overall effectiveness of the framework.

In addition, we further evaluate the impact of the number of retrieved samples ($K$) on {\tool}'s performances, and the results are shown in the Appendix \ref{sec:appendix_retrieved_samples}.

\subsection{Generality and Robustness of {\tool} (RQ3)}

\begin{table}[b]
\centering
\vspace{-0.4cm}
\caption{Performance of {\tool} across two evaluation settings (\%): generalization on unseen memes, and robustness evaluation (original vs. adversarial) on the Hatefulness task. Values in parentheses indicate the performance drop compared to the original memes.}
  
\resizebox{\columnwidth}{!}{
\begin{tabular}{lcc|cccc}
\toprule
\multirow{3}{*}{\textbf{Approach}} & \multicolumn{2}{c|}{\multirow{2}{*}{\textbf{Unseen Memes}}} & \multicolumn{4}{c}{\textbf{Robustness Evaluation}} \\
& \multicolumn{2}{c|}{} & \multicolumn{2}{c}{\textbf{Original Memes}} & \multicolumn{2}{c}{\textbf{Adversarial Memes}} \\
\cmidrule(lr){2-3} \cmidrule(lr){4-5} \cmidrule(lr){6-7}
& Acc & F1 & Acc & F1 & Acc & F1 \\
\midrule
InternVL3-14B & 67.00 & 66.51 & 68.25 & 68.24 & 67.75 $_{(-0.50)}$ & 67.70 $_{(-0.54)}$ \\
+{\tool} & \textbf{72.60} & \textbf{71.78} & \textbf{76.95} & \textbf{75.06} & \textbf{75.54} $_{(-1.41)}$ & \textbf{73.34} $_{(-1.72)}$ \\
\midrule
$\Delta$ (Improvement) & $+5.40$ & $+5.27$ & $+8.70$ & $+6.82$ & $+7.79$ & $+5.64$ \\
\bottomrule
\end{tabular}
\vspace{-0.2cm}
\label{tab:robustness}
}
\end{table}

{To further evaluate the generality and robustness of our proposed framework, we conduct additional experiments on out-of-distribution (unseen) and adversarial memes. We select InternVL3-14B as the backbone model for these evaluations.

\textbf{Generality: Performance on Unseen Memes.} To evaluate whether {\tool} can generalize to entirely new topics without updating its knowledge base, we evaluate it on a newly curated set of 500 recent political memes, crawled from social media platform X spanning from April 1, 2025, to July 1, 2025 (detailed in Appendix), which are strictly excluded from the previous training and testing datasets. As shown in Table~\ref{tab:robustness}, relying solely on the existing misjudgment risk patterns retrieved from the original knowledge base, {\tool} achieves an accuracy of 72.60\% and an F1-score of 71.78\%. This represents a significant improvement of 5.40\% in accuracy and 5.27\% in F1-score compared to the InternVL3-14B model. This substantial gain demonstrates that the misjudgment risk patterns extracted by {\tool} capture the underlying cognitive logic of harm manifestation (e.g., rhetorical devices and implicit analogies) rather than merely memorizing dataset-specific content. Consequently, these misjudgment risk patterns can be effectively migrated to unseen domains and guide the model in detecting these new memes.

\textbf{Robustness: Performance on Adversarial Memes.} To evaluate the robustness of {\tool} against adversarial attacks, we apply the SaltPepper-I-High attack~\cite{aggarwal2023hateproof, mei2025robust}, which injects random white and black pixels across the original memes in a manner that does not compromise the overall perception of semantic content.

Under this high-intensity visual perturbation, {\tool} achieves an accuracy of 75.54\% and an F1-score of 73.34\%. While this reflects a slight sensitivity to extreme noise compared to original images, such behavior can be attributed to its underlying mechanism: our approach relies on extracting fine-grained visual details for structural harm deconstruction, which are naturally more susceptible to severe pixel-level noise than superficial features. Nevertheless, despite this minor degradation, {\tool} still significantly outperforms the vanilla model under the same adversarial conditions, delivering an improvement of 7.79\% in accuracy and 5.64\% in F1-score. This demonstrates that the decision-centric guidance provided by {\tool} remains highly robust and effective even in challenging adversarial environments.}

\section{Discussion}

In this section, we provide several cases to intuitively illustrate the advantages of {\tool} with the misjudgment risk patterns, as well as the human evaluation of the pattern's quality.

\subsection{Case Study}

\begin{figure}[t]

\centering
\includegraphics[width=1.0\linewidth]{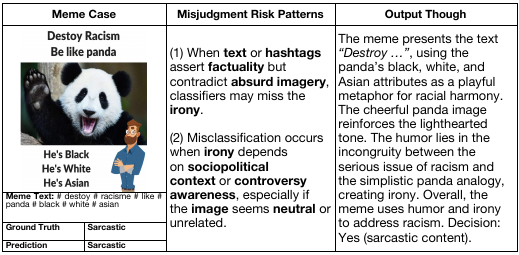}
\vspace{-0.7cm}
\caption{The case study of {\tool}.}
\vspace{-0.7cm}

\label{fig:case_study_one_sample}

\end{figure}

To intuitively illustrate how our framework guides the reasoning process of MLLMs, we present a representative case study in Figure~\ref{fig:case_study_one_sample}. Further analyses of diverse scenarios, such as identifying implicit stereotypes and navigating subjective satire, are provided in the Appendix ~\ref{sec:appendix_case_study}."

\textbf{Accurate Sarcasm Detection through Contextual Incongruity.} 
Figure~\ref{fig:case_study_one_sample} illustrates a meme using an absurd panda analogy for an anti-racist message. To prevent MLLMs from interpreting the lighthearted imagery literally, our framework retrieves critical \textit{misjudgment risk patterns}. These patterns warn that irony is easily missed when ``text contradicts absurd imagery'' and emphasize its dependence on ``sociopolitical context.'' Guided by these insights, the MLLM's reasoning (\textit{Output Thought}) successfully captures the incongruity between the serious racial issue and the visual metaphor, accurately classifying it as ``Sarcastic.'' This case exemplifies how our decision-centric approach proactively mitigates cognitive blind spots. By explicitly injecting risk patterns into the reasoning chain, the framework not only achieves accurate classification but also yields a highly transparent and interpretable decision logic.

\subsection{Human Evaluation}
To validate the quality of the misjudgment risk patterns generated by GPT-4o, we conduct human evaluation by referencing the evaluation framework proposed by~\cite{mazhar2025figurative}.
We randomly sample 20 instances from each of the five tasks within our benchmark, creating an evaluation set of 100 instances in total.
Each meme is associated with a generated misjudgment risk pattern.

We instruct three human evaluators, each with over five years of experience in digital content moderation, to rate these patterns on a 5-point Likert scale (from 1: poor to 5: excellent) based on four key criteria: \textit{relevance}, \textit{correctness}, \textit{actionability}, and \textit{uniqueness}.
The distribution of average scores of these criteria, as illustrated in Figure~\ref{fig:human_eval_results}, indicates that the majority of ratings are concentrated between 4 and 5.
This suggests that the misjudgment risk patterns generated by GPT-4o are well-recognized and align closely with human judgments of potential risks.

\begin{figure}[t]
\centering
\includegraphics[width=\columnwidth]{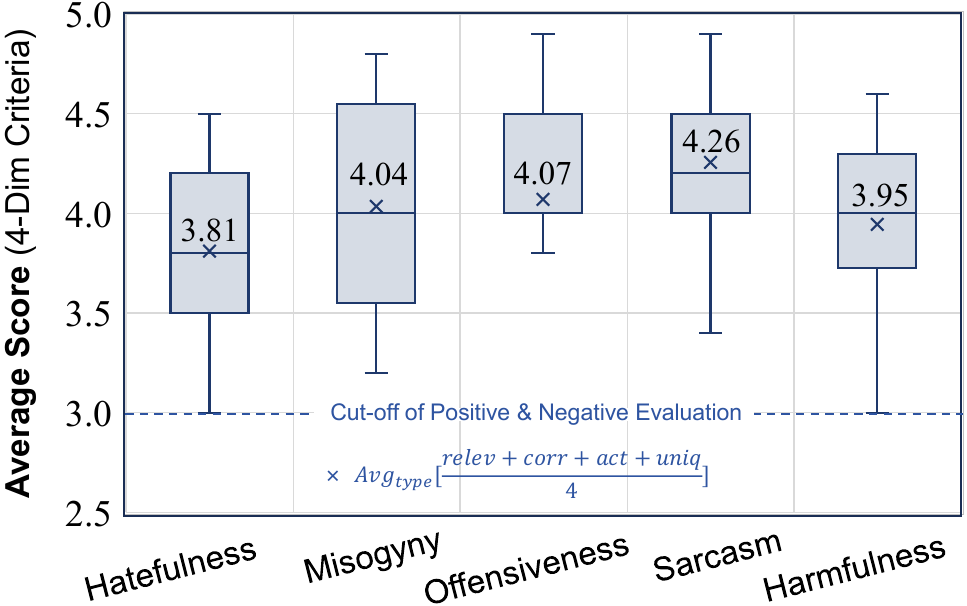}
\vspace{-0.7cm}
\caption{Distribution of human evaluation scores for misjudgment risk patterns generated by GPT-4o.}

\vspace{-0.6cm}

\label{fig:human_eval_results}

\end{figure}

\subsection{Limitations}

\textbf{Inference Latency.} 
{\tool} introduces an additional inference latency of approximately 0.7s per meme compared to the vanilla MLLM. This overhead stems from the retrieval and reranking of risk patterns, as well as processing the augmented context. However, this slight delay is a worthwhile trade-off for the substantial 8.30\% F1-score improvement, particularly in scenarios prioritizing detection accuracy over strict real-time processing.

\textbf{Dependency on MLLM Reliability.} 
The automated pattern elicitation relies on the underlying MLLM, which is susceptible to hallucinations. While we mitigated this in our current knowledge base through human evaluation and manual correction of erroneous patterns, the framework's scalability remains bounded by the MLLM's reliability. Future work could explore leveraging advanced reasoning models or multi-agent consensus frameworks for automated pattern verification, thereby reducing human intervention and enhancing the robustness of the knowledge base.
\section{Conclusion}
In this paper, we propose {\tool}, a cognitive-guided approach that detects harmful memes by learning from and proactively mitigating potential misjudgment risks.
It first constructs a knowledge base, where each meme is deconstructed into a misjudgment risk pattern that captures why and how the MLLM may misinterpret its content.  
For a given target meme, {\tool} retrieves the semantically similar memes, and the relevant patterns are used to dynamically steer the MLLM's reasoning, proactively guiding the MLLM to avoid known misjudgment pitfalls.
Experiments show that {\tool} significantly outperforms the baselines, achieving an average of 8.30\% F1-score improvement, while demonstrating generality and robustness on unseen and adversarial memes.




\bibliographystyle{ACM-Reference-Format}
\bibliography{sample-base}


\appendix

\section{Analysis of Retrieved Samples ($K$)}\label{sec:appendix_retrieved_samples}

\begin{figure}[htbp]
\centering
\includegraphics[width=\columnwidth]{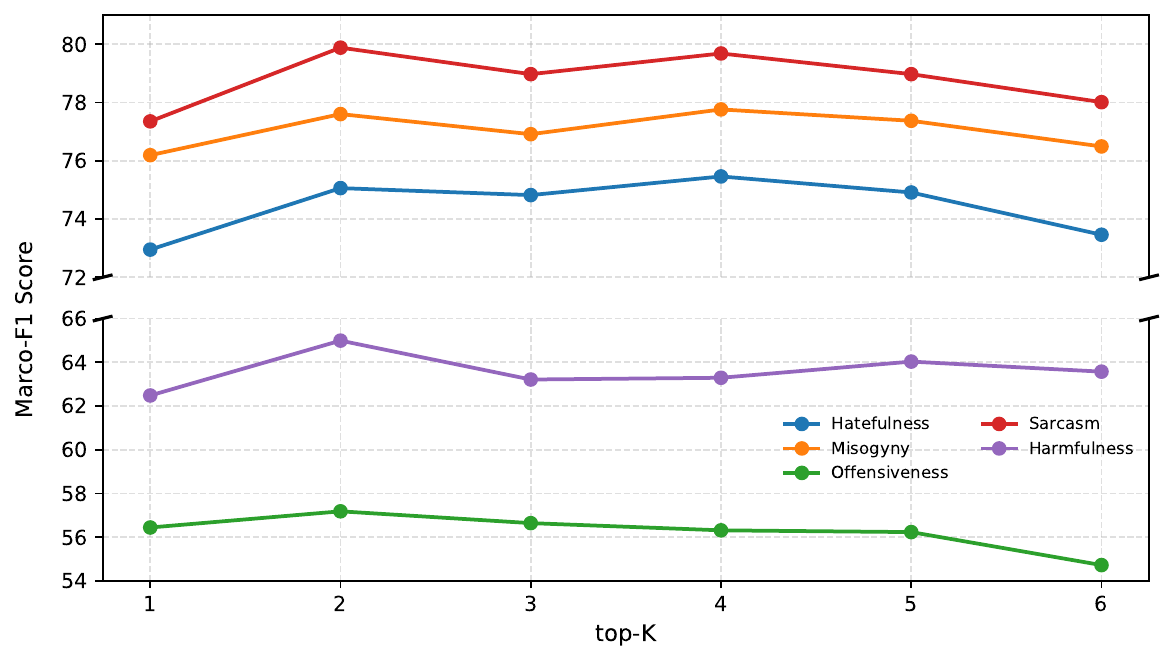}

\caption{Impact of the retrieved case's number ($K$).}

\label{fig:topk_analysis}
\end{figure}

In addition to the ablation study in the Section \ref{sec:ablation_study}, we also investigated the impact of the number of retrieved samples (top-$K$) on model performance, with the results shown in Figure~\ref{fig:topk_analysis}.

As illustrated in the figure, as the value of $K$ increases, the model's performance on most tasks shows a trend of initially rising and then stabilizing or slightly declining. Specifically, the model achieves optimal or near-optimal performance on most tasks when $K$ is between 2 and 4. This suggests that providing a small number of high-quality misjudgment risk patterns from similar memes can effectively guide the model's reasoning. However, when $K$ becomes too large (e.g., greater than 4), introducing too much information may introduce noise that interferes with the model's judgment. Considering both performance and efficiency, we set the optimal value of $K$ to 2 in our experiments.

\section{Pseudocode for {\tool}}
In this section, we have provided the pseudocode of the pattern retrieval, so as to improve the reproducibility.

\begin{algorithm}[htbp]
\small
	\caption{Risk-aware Pattern Retrieval.
    } 
 \label{alg:auto_prompting}
	\KwIn{Test sample $m_{test}$, Knowledge base $K$.}
	\KwOut{Final prediction $y_{pred}$ and rationale $\text{analysis}$.}
    \SetKwProg{Fn}{Function}{}{}{end}
    \Fn{RetrieveAndDetect($m_{test}, K$):}{
        $v_{test}^{\text{img}}, v_{test}^{\text{txt}} \gets E(\text{img}_{test}, \text{txt}_{test})$\;
        $\text{Candidates}_b \gets \textproc{Retrieve}(K, \text{'benign'}, v_{test}^{\text{img}}, v_{test}^{\text{txt}}, N)$\;
        $\text{Candidates}_h \gets \textproc{Retrieve}(K, \text{'harmful'}, v_{test}^{\text{img}}, v_{test}^{\text{txt}}, N)$\;
        $\text{CandidatePool} \gets \text{Candidates}_b \cup \text{Candidates}_h$\;
        
        $\text{ScoredList} \gets []$\;
        \For{each candidate $c$ in $\text{CandidatePool}$}{
            $S_{\text{img}} \gets R_{img}(\text{img}_{test}, c.\text{img})$\;
            $S_{\text{txt}} \gets R_{txt}(\text{txt}_{test}, c.\text{txt})$\;
            
            $S_{\mathcal{T}} \gets \textproc{MatchKnowledgeTree}(m_{test}.\text{tree}, c.\text{tree})$\;
            
            $S_{\text{final}} \gets \alpha \cdot S_{\text{img}} + (1 - \alpha) \cdot S_{\text{txt}} + S_{\mathcal{T}}$\;
            Append $(c, S_{\text{final}})$ to $\text{ScoredList}$\;}
        Sort $\text{ScoredList}$ in descending order by $S_{\text{final}}$\;
        $\text{TopK\_Cases} \gets$ Select top $K$ items from $\text{ScoredList}$\;
        
        $P_{Hateful} \gets \textproc{ExtractPattern}(\text{TopK\_Cases}, \text{`hateful'})$\;
        $P_{Benign} \gets \textproc{ExtractPattern}(\text{TopK\_Cases}, \text{`benign'})$\;
        $\text{prompt} \gets \textproc{ConstructPrompt}(m_{test}, P_{Hateful}, P_{Benign})$\;
        $(\text{analysis}, y_{pred}) \gets \text{MLLM}_{\text{judge}}.\textproc{Query}(\text{prompt})$\;
    }
    \Return \text{analysis}, $y_{pred}$
\end{algorithm}

\section{Prompt for Harmful Meme Detection}\label{sec:appendix_prompt}

To improve the reproducibility of {\tool}, we provide the contents of three prompts that are used in the harmful meme detection:

\begin{itemize}[leftmargin=*]
    \item \textbf{Figure \ref{fig:prompt_tree}'s Prompt:} Hierarchical harm tree construction (Step-1 of {\tool}), where the prompt is used in Section~\ref{sec:hier_harm_deconstruction} hierarchical harm deconstruction.
    \item \textbf{Figure \ref{fig:prompt_detection}'s Prompt:} Harmful meme detection (Step-3 of {\tool}), where the prompt is used in Section~\ref{sec:prompt_construction} pattern-augmented prompt construction.
    \item \textbf{Figure \ref{fig:prompt_pattern}'s Prompt:} Misjudgment risk pattern generation (Step-1 of {\tool}), where the prompt is used in Section~\ref{sec:causal_misjudg_att} causal misjudgment attribution.
\end{itemize}

\section{Case Study}
\label{sec:appendix_case_study}

To more intuitively understand our framework's reasoning process, we first analyze several correctly predicted cases. These cases demonstrate the misjudgment risk patterns retrieved by our approach and show the key content of the reasoning process. We also conduct a failure case analysis to explore the current limitations and future directions of our approach, as shown in Figure~\ref{fig:case_study}.

\textbf{Detecting Implicit Harm Masked by Neutral Language and Humor.}
A significant challenge in meme detection is identifying harmful intent hidden behind seemingly innocuous text or humor. Our approach excels in this area by retrieving relevant risk patterns that guide MLLM to look beyond the surface. In \textbf{Sample (a)}, the meme uses a common phrase, ``spot the difference.'' Our approach, guided by the retrieved misjudgment risk pattern warning against \textbf{``neutral text paired with images that implicitly dehumanize,''} correctly identifies the racist comparison. The output reasoning clearly states that ``the juxtaposition of monkeys and children carries an implicit risk of dehumanization,'' leading to the correct ``Hateful'' prediction. Similarly, in \textbf{Sample (b)}, the meme employs irony to convey a misogynistic message. Our approach retrieves a pattern highlighting that \textbf{``humor or irony can obscure misogyny.''} This prompts a deeper analysis, as reflected in the output: MLLM correctly discerns that the meme ``trivializes women's rights by equating equality with physical confrontation,'' thus identifying its underlying misogynistic nature.

\textbf{Accurate Sarcasm Detection through Contextual Incongruity.}
Our approach's ability to analyze context extends to non-harmful categories like sarcasm, where understanding the intent is key. The task for \textbf{Sample (c)} is to identify if the content is sarcastic. The meme uses an absurd analogy of a panda to deliver an anti-racist message. Our system retrieves risk patterns related to \textbf{``irony depends on sociopolitical context''} and situations where \textbf{``text...contradicts absurd imagery.''} Guided by these insights, MLLM correctly focuses on the incongruity between the serious issue of racism and the simplistic panda analogy. This understanding allows it to correctly classify the meme as ``Sarcastic,'' demonstrating the approach's ability to interpret nuanced forms of expression.

\textbf{Failure Analysis: The Subjective Boundary Between Satire and Offense.}
In \textbf{Sample (d)}, a political meme satirizing Donald Trump is labeled ``Offensive,'' but our model predicted ``Benign.'' The analysis reveals a key challenge: while the approach successfully retrieved the relevant risk pattern--- \textbf{``When offensive content is framed through satire...it may be overlooked''}---it struggled with the final judgment. It over-prioritized the satirical context and the targeting of a public figure, leading to an incorrect classification. This highlights the difficulty in navigating the subjective boundary between sharp satire and offensive content, marking a clear area for future improvement.

In summary, these case studies demonstrate that our misjudgment driven approach enhances a model's ability to decipher the implicit and contextual cues in memes. This leads to a more nuanced understanding and accurate classification across different categories, from hate speech to sarcasm.

\begin{figure*}[t]

\centering
\includegraphics[width=1.0\linewidth]{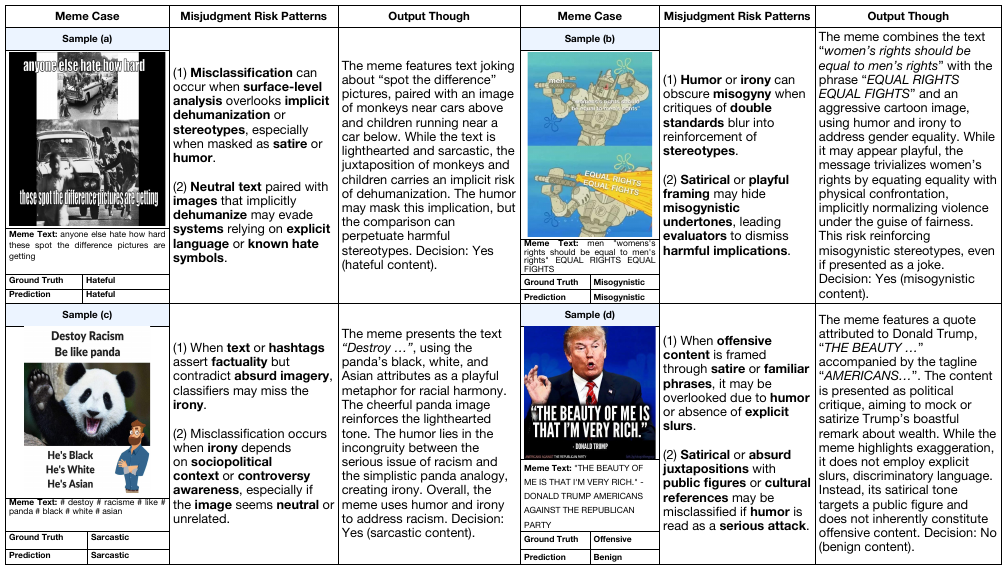}
\caption{The case study of {\tool} on four examples.}
\label{fig:case_study}
\end{figure*}

\section{Details of the Unseen Meme Dataset}
\label{sec:appendix_unseen_dataset}
To rigorously evaluate the out-of-distribution (OOD) generalization capability of \tool (as discussed in RQ3), we curated a new dataset comprising 500 recent political memes. This section details the data collection, annotation process, and statistical distribution of this unseen dataset.

\paragraph{Data Collection and Thematic Scope.} 
The dataset was sourced from the social media platform X (formerly Twitter) spanning from April 1, 2025, to July 1, 2025. This specific timeframe ensures that the collected memes encapsulate recent political events and emerging rhetorical trends that are strictly absent from our primary training and testing sets. To comprehensively evaluate the model's robustness across diverse contexts, the curated memes cover a wide array of highly debated topics, including: (1) U.S. partisan politics and macro-ideologies; (2) foreign policy and international conflicts; (3) gender issues, LGBTQ+ rights, and family values; (4) gun rights and gun control; and (5) government credibility, economic policies, and social welfare.

\paragraph{Annotation Process.} 
The entire annotation process was strictly guided by a unified definition of harmful content~\cite{lin2025ask}: \textit{``speech or material that mocks or ridicules a targeted person or organization, or has the potential to cause emotional discomfort to any individual, politician, celebrity, or the general public.''} Based on this explicit criterion, we adopted a robust model-in-the-loop annotation pipeline to balance efficiency and accuracy. Initially, an ensemble of multiple SOTA MLLMs was prompted with this definition to generate preliminary labels for each meme, followed by a majority voting mechanism to determine the consensus prediction. Subsequently, a rigorous manual review phase was conducted by human experts. These annotators verified the labels generated by the models against the predefined criteria, corrected any misclassifications (particularly for memes involving implicit satire or complex cultural contexts), and finalized the categorization.

\paragraph{Dataset Statistics.} 
The final dataset consists of 500 annotated memes, comprising 300 (60\%) ``Benign'' samples and 200 (40\%) ``Harmful'' samples (see Figure \ref{fig:unseen_sample} for examples). We utilize this dataset to evaluate whether {\tool} can effectively transfer its learned misjudgment risk patterns to novel, unseen political domains, rather than merely relying on prior knowledge specific to the dataset.

\begin{figure*}[t]

\centering
\includegraphics[width=1.0\linewidth]{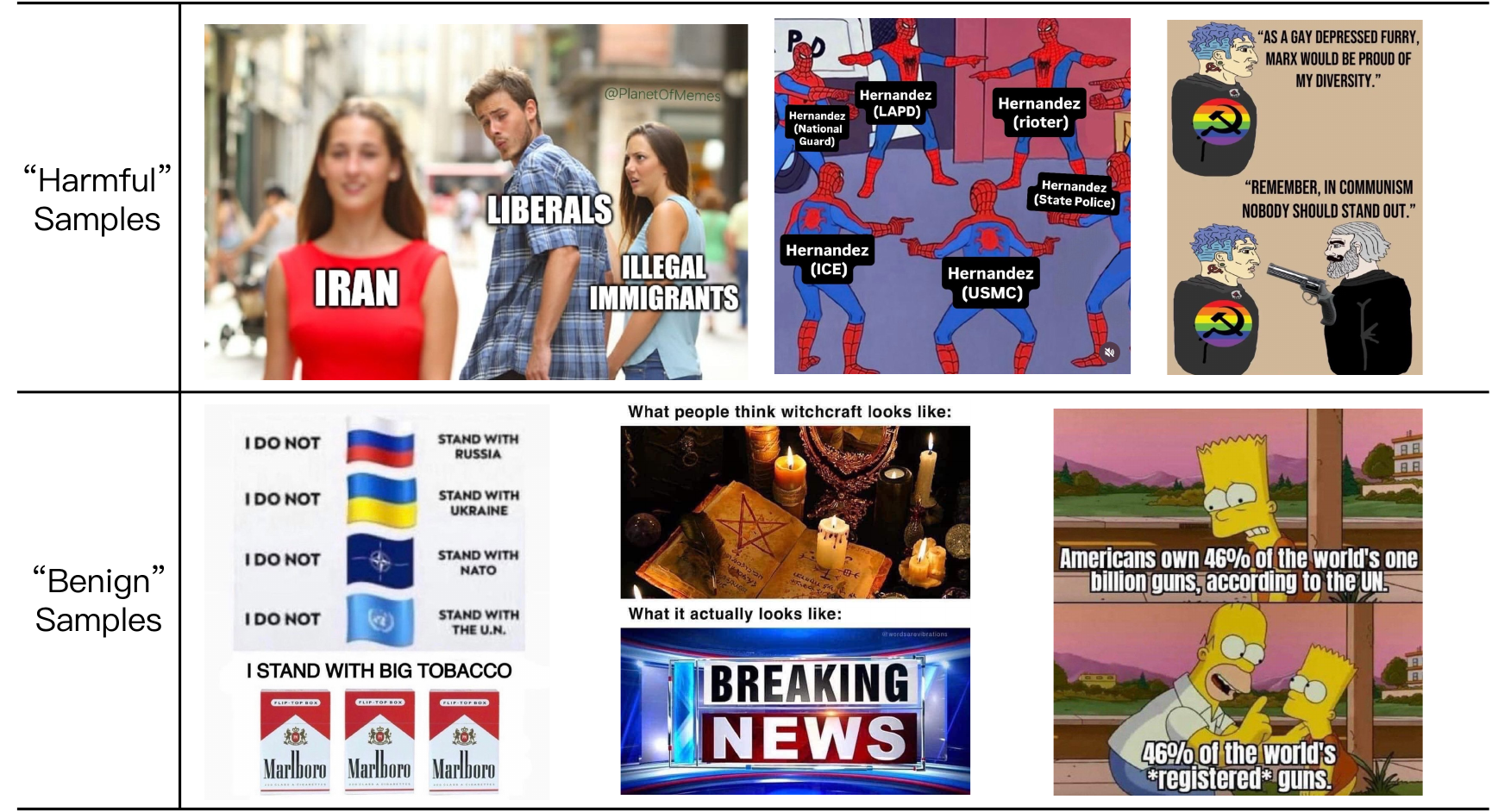}
\caption{Examples of unseen memes.}

\label{fig:unseen_sample}
\end{figure*}

\begin{figure*}[t]
\centering
\includegraphics[width=\linewidth]{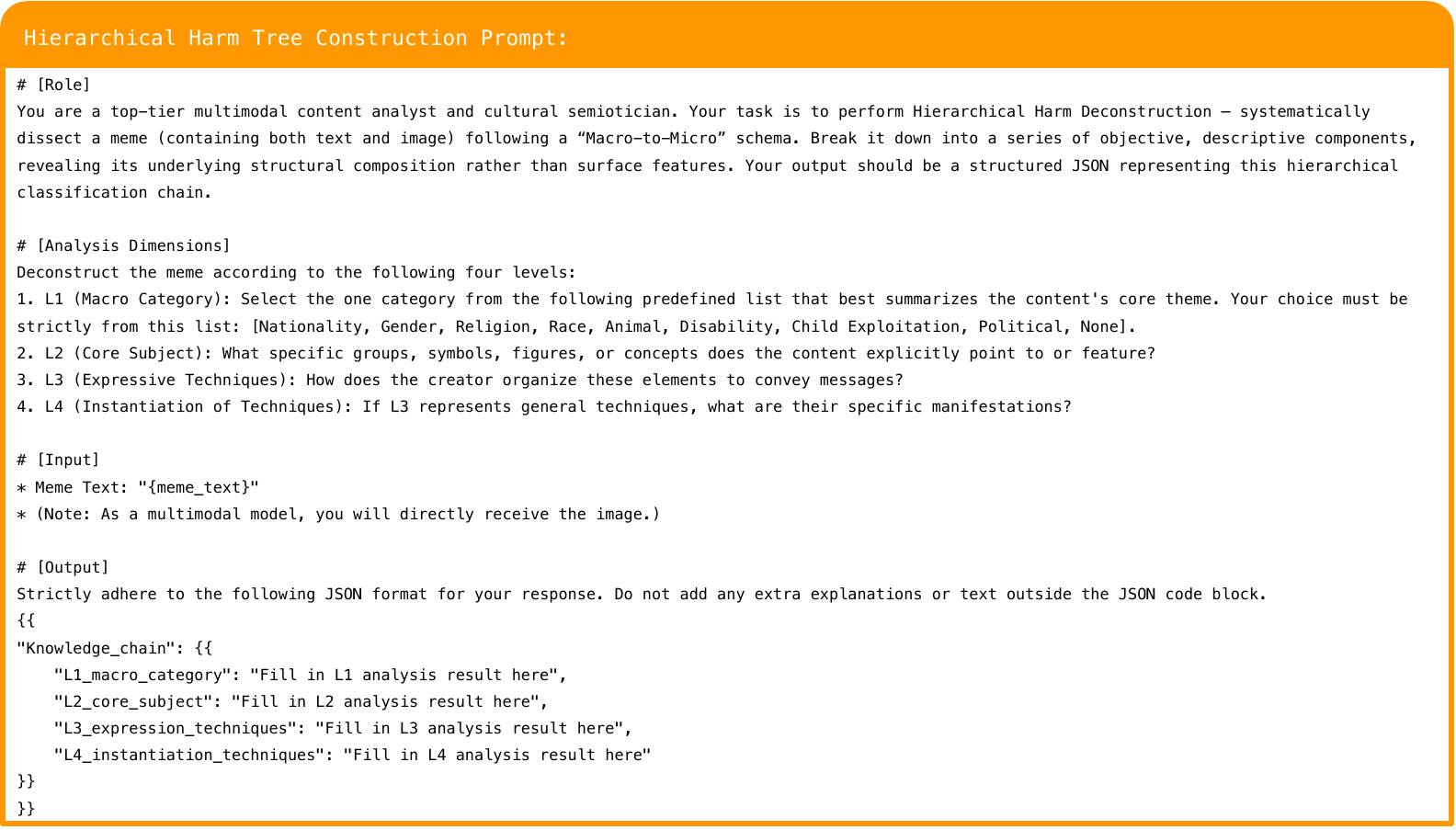}
\caption{Prompt for hierarchical harm tree construction.}
\label{fig:prompt_tree}
\end{figure*}

\begin{figure*}[t]
\centering
\includegraphics[width=\linewidth]{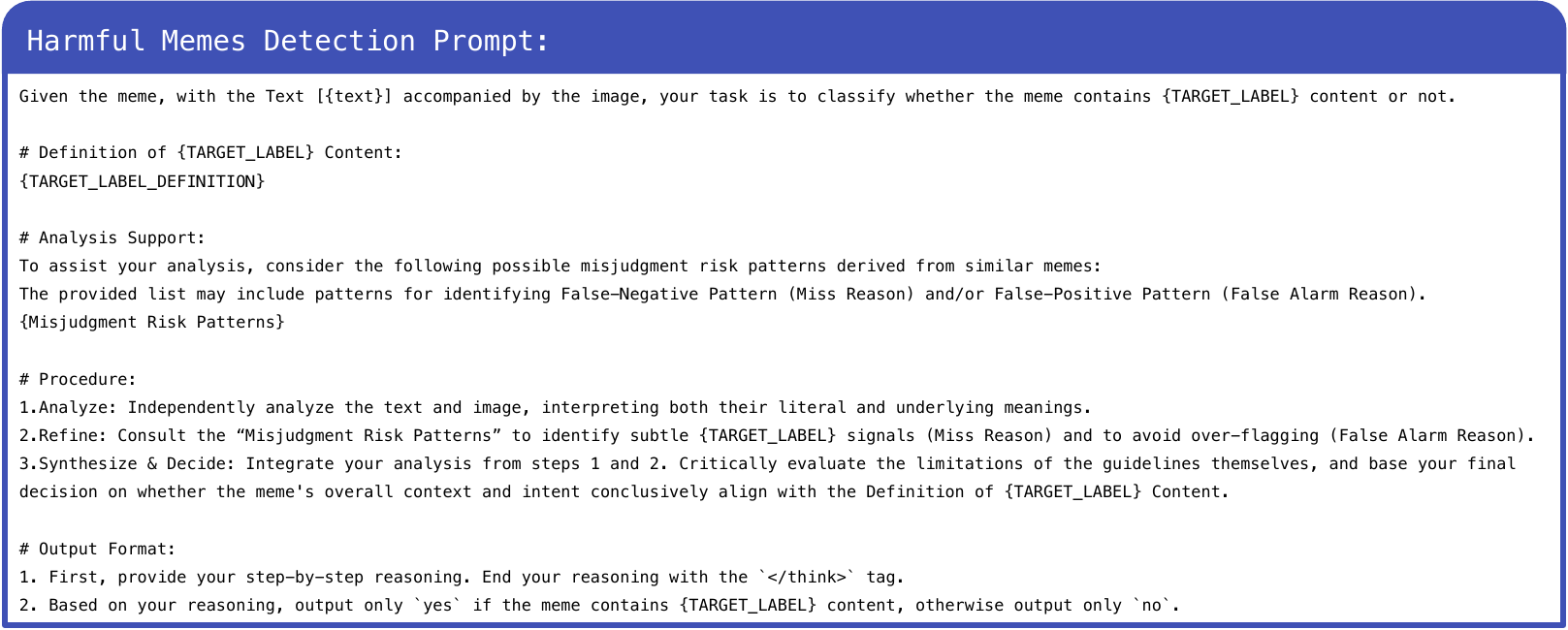}
\caption{Prompt for harmful memes detection.}
\label{fig:prompt_detection}
\end{figure*}

\begin{figure*}[t]
\centering
\includegraphics[width=\linewidth]{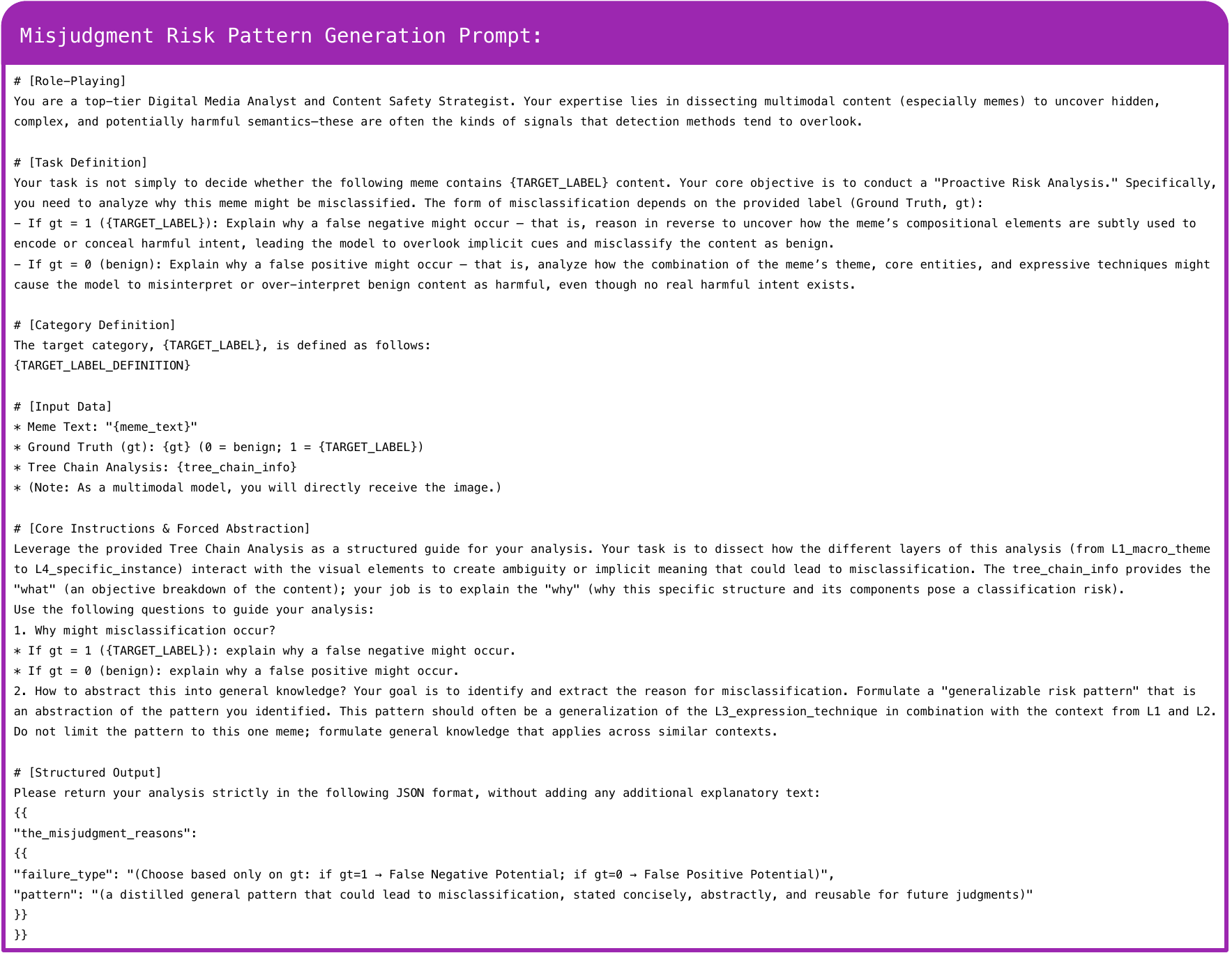}
\caption{Prompt for misjudgment risk pattern generation.}
\label{fig:prompt_pattern}
\end{figure*}

\end{document}